\documentclass[a4paper,fleqn]{cas-dc}

\usepackage[authoryear,longnamesfirst]{natbib}
\usepackage{lineno,hyperref}
\usepackage{ulem}
\usepackage{algorithm2e,setspace}
\usepackage{subfigure}
\usepackage{multirow}
\usepackage{bbding}
\usepackage{mathrsfs}
\usepackage{amsthm,amsmath,amssymb}
\usepackage{tabularx}
\usepackage{natbib}
\usepackage{booktabs}

\def\tsc#1{\csdef{#1}{\textsc{\lowercase{#1}}\xspace}}
\tsc{WGM}
\tsc{QE}
\tsc{EP}
\tsc{PMS}
\tsc{BEC}
\tsc{DE}

\begin{document}
\let\WriteBookmarks\relax
\def\floatpagepagefraction{1}
\def\textpagefraction{.001}

\shorttitle{}

\shortauthors{M. Yao et~al.}

\title [mode = title]{FMRFT: Fusion Mamba and DETR for Query Time Sequence Intersection Fish Tracking}



%

\author[1,2,3]{Mingyuan Yao}[
  style=chinese,
  orcid=0009-0007-4882-2985
]



\ead{yaomingyuan@cau.edu.cn}


\credit{Conceptualization, Methodology, Software, Validation, Formal analysis, Investigation, Data Curation, Writing - Original Draft, Writing - Review \& Editing, Visualization}

\author[1,2,3]{Yukang Huo}[
  style=chinese,
  orcid=0009-0001-9569-7028
]

\ead{huoyukang@cau.edu.cn}

\credit{Conceptualization, Methodology, Validation, Formal analysis, Data Curation, Writing - Review \& Editing, Visualization}

\author[1,2,3]{Qingbin Tian}[
  style=chinese,
  orcid=0009-0004-1106-5560
]

\ead{tianqingbin@cau.edu.cn}

\credit{Methodology, Data Curation, Writing - Review \& Editing}

\author[1,2,3]{Jiayin Zhao}[
  style=chinese,
  orcid=0009-0008-3244-2635
]

\ead{zhaojiayin@cau.edu.cn}

\credit{Methodology, Data Curation, Writing - Review \& Editing}

\author[1,2,3]{Xiao Liu}[
  style=chinese,
  orcid=0009-0004-5727-1239
]

\ead{lxiao@cau.edu.cn}
\credit{Methodology, Data Curation, Writing - Review \& Editing}

\author[4]{Ruifeng Wang}[
  style=chinese,
  orcid=0009-0004-9051-0094
]

\ead{sweefongreggiewong@cau.edu.cn}

\credit{Writing - Review \& Editing}

\author[5]{Lin Xue}[
  style=chinese,
]

\ead{xxtxuelin@buu.edu.cn}

\credit{Validation, Writing - Review \& Editing}

\author[1,2,3]{Haihua Wang}[
  style=chinese,
  orcid=0000-0002-7415-4753
]


\ead{whaihua@cau.edu.cn}
\cormark[1]

\credit{Resources, Writing - Review \& Editing, Supervision, Project administration, Funding acquisition}

\affiliation[1]{organization={National Innovation Center for Digital Fishery},
  addressline={No. 17, Qinghua East Road, Haidian District},
  city={Beijing},
  postcode={100083},
  country={China}}

\affiliation[2]{organization={Key Laboratory of Smart Farming Technologies for Aquatic Animal and Livestock, Ministry of Agriculture and Rural Affairs},
  addressline={No. 17, Qinghua East Road, Haidian District},
  city={Beijing},
  postcode={100083},
  country={China}}

\affiliation[3]{organization={College of Information and Electrical Engineering, China Agricultural University},
  addressline={No. 17, Qinghua East Road, Haidian District},
  city={Beijing},
  postcode={10083},
  country={China}}

\affiliation[4]{organization={College of Engineering, China Agricultural University},
  addressline={No. 17, Qinghua East Road, Haidian District},
  city={Beijing},
  postcode={10083},
  country={China}}

\affiliation[5]{organization={Smart City College, Beijing Union University},
  addressline={No. 97, North Fourth Ring East Road, Chaoyang District},
  city={Beijing},
  postcode={100101},
  country={China}}

\cortext[cor1]{Corresponding author}



\begin{abstract}
  Early detection of abnormal fish behavior caused by disease or hunger can be achieved through fish tracking using deep learning techniques, which holds significant value for industrial aquaculture. However, underwater reflections and some reasons with fish, such as the high similarity, rapid swimming caused by stimuli and mutual occlusion bring challenges to multi-target tracking of fish. To address these challenges, this paper establishes a complex multi-scenario sturgeon tracking dataset and introduces the FMRFT model, a real-time end-to-end fish tracking solution.  The model incorporates the low video memory consumption Mamba In Mamba (MIM) architecture, which facilitates multi-frame temporal memory and feature extraction, thereby addressing the challenges to track multiple fish across frames. Additionally, the FMRFT model with the Query Time Sequence Intersection (QTSI) module effectively manages occluded objects and reduces redundant tracking frames using the superior feature interaction and prior frame processing capabilities of RT-DETR. This combination significantly enhances the accuracy and stability of fish tracking. Trained and tested on the dataset, the model achieves an $IDF_1$ score of 90.3\% and a $MOTA$ accuracy of 94.3\%. Experimental results show that the proposed FMRFT model effectively addresses the challenges of high similarity and mutual occlusion in fish populations, enabling accurate tracking in factory farming environments.

\end{abstract}





\maketitle

\section{Introduction}
With the rapid development of aquaculture, real-time detection and assessment of fish conditions play a crucial role in improving fish farming efficiency and enhancing management practices \citep{paper01}. Monitoring fish conditions can facilitate the timely detection of feeding issues in aquaculture to achieve precise feeding, which reduces feed waste, minimizes water pollution, and enhances yields\citep{paper02}. Fish exhibit various behavioral responses to external stimuli such as light, water quality, and breeding density. By tracking fish status, the aquaculture system can gain valuable insights into fish health, environmental adaptation, and more \citep{paper03,paper04}. Compared to traditional methods based on sensors or manual observation, fish tracking with computer vision offers advantages such as real-time monitoring, non-contact observation, and non-interference. This approach is an effective means of achieving intgent management in large-scale aquaculture operations \citep{paper05}.

Multi-target tracking is a computer vision task focused on localizing and tracking multiple targets within a video sequence. A variety of algorithms have emerged in this field, with the two most dominant strategies being Tracking by Detection and Tracking by Query. The core idea of Tracking by Detection is to first use a target detection algorithm to identify targets within each frame of the video. Subsequently, matching algorithms such as DeepSort \citep{paper06}, ByteTrack \citep{paper07}, or other related algorithms are employed to associate and match the detected targets across consecutive frames, thereby enabling the tracking of the target trajectory. This method lies in its reliance on a powerful target detector that provides accurate target position information. However, it still encounters significant challenges in complex scenes involving occlusion, lighting changes, and fast target motion. In contrast, Tracking by Query is an emerging approach that represents each target as a query(typically a feature vector) to search and match targets throughout a video sequence. Algorithms such as TransMOT \citep{paper08}, TransCenter \citep{paper09}, and MOTR \citep{paper10} have been developed under this paradigm. Query-based tracking methods specifically address challenges such as variations in target appearance and occlusion. These approaches demonstrate greater robustness in complex environments by leveraging the continuity of the target's features.

\begin{figure}
  \centering
  \includegraphics[width=0.9\linewidth]{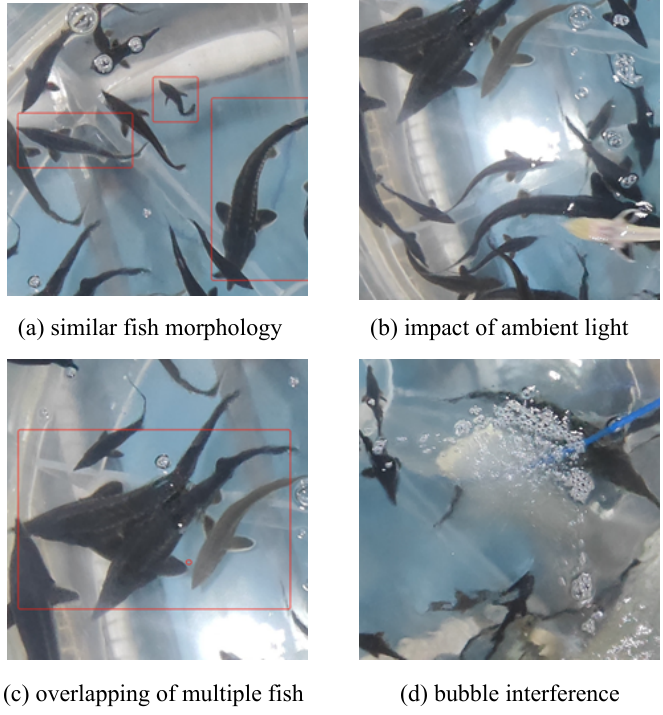}
  \caption{Multi-Target Fish Tracking Barriers.}
  \label{fig1}
  \vspace{-5px}
\end{figure}

However, fish tracking often presents greater complexities compared to traditional object tracking scenarios. Firstly, as shown in Figure~\ref{fig1}(a), the morphological changes of individual fish are not pronounced at different growth stages or during the same period due to behaviors such as respiration and swimming, which increases the complexity of target identification. Secondly, as illustrated in Figure~\ref{fig1}(b), varying lighting conditions at different angles within the fish tank, along with light refraction, scattering, and absorption, result in low image contrast and clarity. These factors significantly hinder the accurate detection and tracking of fish targets. As shown in Figure~\ref{fig1}(c, d), fish frequently cover each other while swimming, particularly in high-density aquaculture environments, where such occlusions pose significant disruptions to continuous target tracking. Additionally, bubbles generated by oxygenators, feed residues in the water, and sensor equipment may have similar texture or brightness characteristics to the fish targets in the image, further complicating fish tracking.

In response to these challenges, this paper proposes a real-time fish tracking model, FMRFT, designed for tracking fish in complex factory farming scenarios. The main contributions of this paper are as follows:
\begin{enumerate}[(1)]
  \item Innovative Framework Fusion: We have innovatively fused Mamba In Mamba (MIM) and the RT-DETR within the existing MOTR framework to achieve accurate tracking of fish targets. This fusion strategy enhances the model's ability to effectively address occlusion and similarity of fish in complex environments.
  \item Novel Query Time Sequence Intersection: We propose a novel query interaction module (QTSI), which facilitates the interaction and fusion of information by calculating the Intersection Over Union (IOU) between Tracking Query, Detect Query, and real frames during the training phase. This design significantly reduces reliance on a single Tracking Query and effectively prevents the generation of multiple redundant detection frames for a single target.
  \item Enhanced Data Fusion Method: To further improve the depth and breadth of feature extraction, we designed a new data fusion method, Fusion MIM, which deeply fuses MIM feature information at different scales and strengthens the model's ability to extract features at multiple levels through feature interaction.
  \item Innovative Temporal Tracking Query Interaction Module : We introduced a Mamba Query Interaction Module (MQIM) that enables tracking queries to be learned through deeper interaction with Decoder layer outputs. This interaction mechanism enhances the model's adaptability to target changes in dynamic scenes and improves tracking stability.
  \item New Multi-Fish Tracking Dataset: A new multi-objective fish tracking dataset has been established, covering sturgeon fish tracking data from various culture scenarios and containing a total of 8,000 high-quality sturgeon fish tracking images. This dataset provides a valuable visual resource for fish behavior analysis and health assessment.
\end{enumerate}

The main contents of the remaining chapters are as follows: Section 2 reviews previous work in fish tracking, and provides a brief introduction to the Mamba and DETR modules. Section 3 details the proposed FMRFT method. Section 4 presents comparative and ablation experiments, along with visualizations of the experimental results. Finally, Section 5 concludes the paper and suggests directions for future research.

\section{Related Work}
Multi-Object Tracking (MOT) is a key technology in the field of computer vision, which widely applied in areas such as autonomous driving, intelligent surveillance, and behavior recognition. Multi-object tracking, despite advances in image processing, continues to face significant challenges. Common issues such as occlusion, object deformation, motion blur, crowded environments, rapid movement, changes in illumination, and variations in scale—challenges that also impact single-target tracking—persist in this domain. Moreover, multi-object tracking introduces additional complexities, including the need for precise trajectory initialization and termination, as well as managing mutual interference between visually similar targets. These challenges make multi-target tracking a particularly demanding area of research, warranting sustained attention and innovation within the field. Currently, two primary strategies dominate multi-object tracking methodologies, both of which are detailed below:

\subsection{Tracking by Detection}
\begin{figure}
  \centering
  \includegraphics[width=0.9\linewidth]{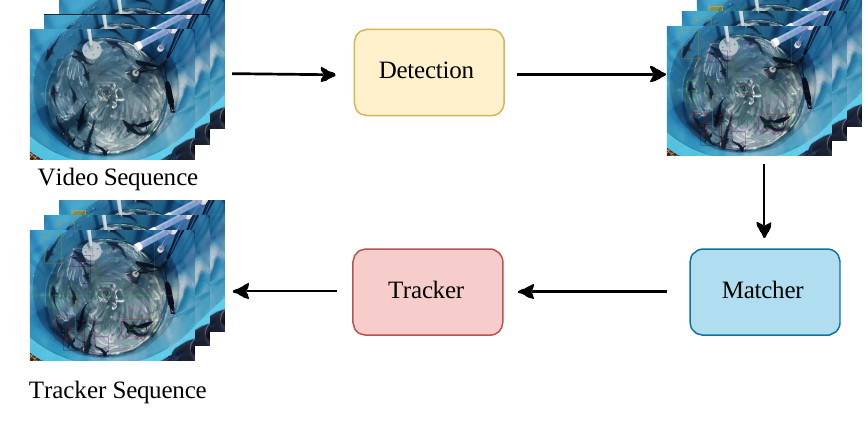}
  \caption{Tracking by Detection Structure.}
  \label{fig2}
  \vspace{-5px}
\end{figure}

Detection-based tracking is a widely adopted paradigm in the field of MOT\citep{paper11}, as illustrated in Figure~\ref{fig2}. This approach is typically divided into two key steps: target detection and target association\citep{paper12}. In the detection phase, various deep learning models are employed to identify objects of interest\citep{paper12}. However, the primary challenge lies in the target association step, which involves maintaining the trajectory of each object across frames. Several well-established methods are employed for this purpose, including Linear Regression\citep{paper14}, Mean Drift\citep{paper15}, Hidden Markov Models\citep{paper16}, the Kalman Filter\citep{paper17,paper18}, the Extended Kalman Filter\citep{paper19}, and the Particle Filter\citep{paper20,paper21,paper22,paper23,paper24,paper25,paper26,paper27}. Each of these methods offers different strengths depending on the specific tracking conditions and the complexity of the environment.

As the originator of detection-based target tracking, Bewley et al. \citep{paper13} developed the first efficient online multi-target tracking method, SORT, which achieves fast and accurate tracking of multiple targets in a video by using a Kalman filter for motion prediction and a Hungarian algorithm for data association. Wojke et al. \citep{paper06} proposed the DeepSORT algorithm, building upon SORT. By introducing appearance features obtained through deep learning feature extraction, DeepSORT addresses the identity-switching problem that SORT encounters when dealing with occlusion and long-term tracking, thereby enhancing the robustness of multi-target tracking. StrongSORT \citep{paper28} further improves multi-target tracking by enhancing object detection, feature embedding, and trajectory association, as well as introducing the AFLink and GSI algorithms. By associating nearly all detection frames instead of just high-scoring ones, StrongSORT improves tracking performance and accuracy. ByteTrack \citep{paper07} addresses the issues of occlusion and low-scoring detection frames, which may cause SORT to miss real objects and fragment trajectories, by associating almost all detection frames instead of just high-scoring ones. The OC-SORT \citep{paper29} algorithm enhances the traditional SORT method by introducing the Observation-centre Re-Update and Observation Centre Momentum, which solves problems related to cumulative error and inaccurate direction estimation in cases of occlusion and nonlinear motion.

In detection-based fish tracking approaches, mainstream target detection algorithms are typically employed to detect fish. For instance, Martija et al. \citep{paper30} implemented an end-to-end tracking and detection algorithm by redesigning the Deep Hungarian Network to compute discriminative affinity scores for predictive detection between consecutive frames. This was combined with the Faster R-CNN model to detect fish in field-captured video sequences for multi-fish tracking. Sun et al. \citep{paper31} proposed a tracking technique based on the YOLOv7-DCN and SORT algorithms, which tracks the primary targets in fishing vessel operations by employing enhanced target detection and counting algorithms that integrate Kalman filters and Hungarian algorithms. Wang et al. \citep{paper32} developed a parallel shape index feature-based fish tracking algorithm, which detects the head and center of the fish body and integrates the SORT framework and Kalman filter to accurately track the movement trajectories of large numbers of zebrafish. Gong et al. \citep{paper33} achieved efficient and accurate underwater fish tracking by incorporating the CBAM attention mechanism into the YOLOv4-tiny model to enhance feature learning, in combination with a SORT tracker.

\subsection{Tracking by Query}
\begin{figure}
  \centering
  \includegraphics[width=0.9\linewidth]{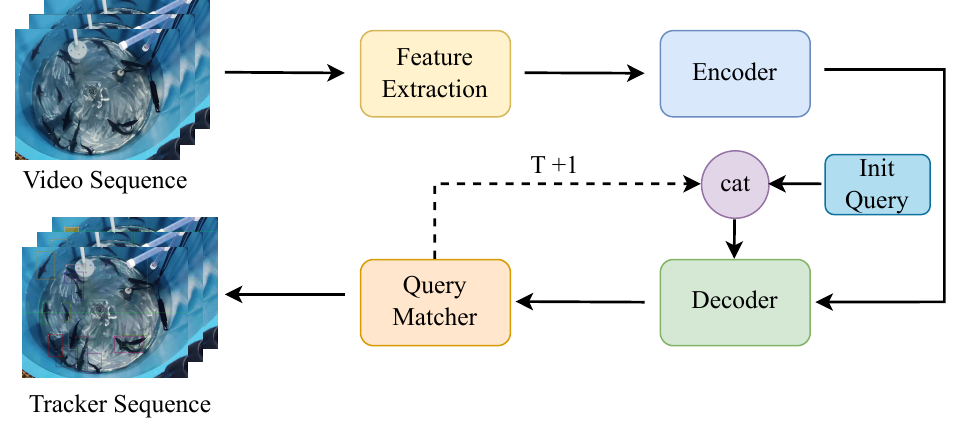}
  \caption{Tracking by Query using TransTrack Formation.}
  \label{fig3}
  \vspace{-5px}
\end{figure}

\begin{figure}
  \centering
  \includegraphics[width=0.9\linewidth]{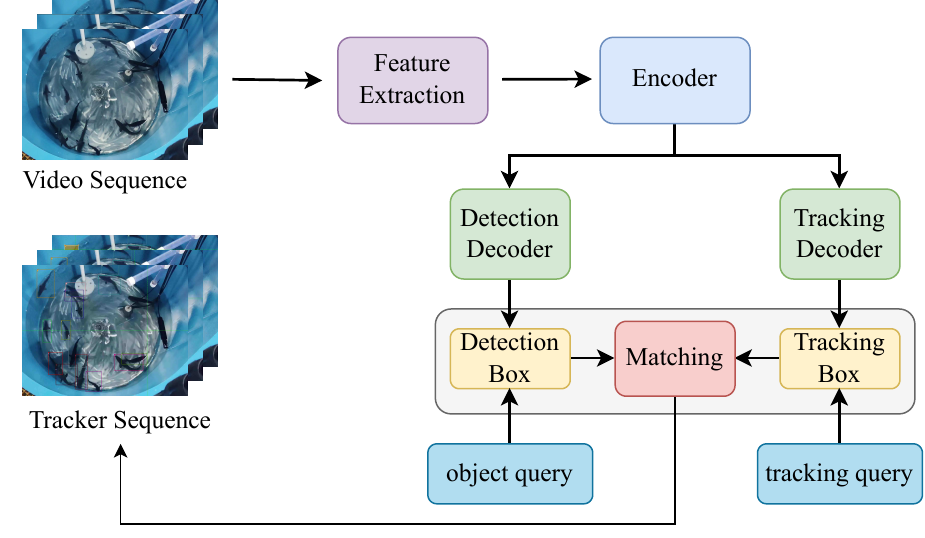}
  \caption{Tracking by Query using TrackFormer Formation.}
  \label{fig4}
  \vspace{-5px}
\end{figure}

With the widespread application of attention mechanisms, particularly Transformers, in the field of computer vision, query-guided target tracking methods have demonstrated significant advantages in tracking robustness, thereby offering new research perspectives and breakthroughs in MOT\citep{paper34}. Currently, TransTrack \citep{paper35} and TrackFormer \citep{paper36} are two representative tracking frameworks that utilize the Transformer architecture to address MOT tasks. TransTrack integrates target detection and association into a unified framework by utilizing the Transformer's attention mechanism, as depicted in Figure~\ref{fig3}. In contract, the TrackFormer algorithm introduces a Transformer-based encoder-decoder architecture and employs an autoregressive trajectory query mechanism, effectively addressing key challenges in multi-target tracking, including data association, identity preservation, and spatiotemporal trajectory prediction. Meanwhile, it also enables end-to-end trainable multi-target tracking and segmentation, as shown in Figure~\ref{fig4}.

Additionally, Xu et al. \citep{paper09} proposed a Transformer-based multi-object tracking architecture that, for the first time, addresses the challenges  of under-detection and computational inefficiency in crowded scenarios by introducing dense image correlation detection queries and efficient sparse tracking queries. The MOTR \citep{paper10} algorithm further enhances accuracy and efficiency in video sequence tracking by introducing a "track query" and iterative prediction mechanism. This approach improves the post-processing stage, that traditionally relies on heuristic correlations based on motion and appearance similarity, and addresses the challenge of exploiting temporal variations in video sequences in an end-to-end manner. Furthermore, the MOTRv2 \citep{paper37} algorithm enhances the detection performance of MOTR by integrating a pre-trained YOLOX object detector to generate proposals as anchors, significantly boosting both the accuracy and efficiency of multi-target tracking.

In recent years, the flexibility of the Query-Key mechanism in Transformer architectures has led numerous researchers to adopt this model for multi-fish tracking applications. Gupta et al. \citep{paper38} proposed a deep fish tracking network named DFTNet, which combines a twin network for encoding appearance similarity and an Attention Long Short-Term Memory network for capturing motion similarity between consecutive frames, thereby enabling efficient fish tracking. Li et al. \citep{paper39} introduced a Transformer-based multi-fish tracking model, incorporating a Multiple Association approach to enhance tracking fault tolerance by integrating simple cross-linking matches in the Identification matching module. Liu et al. \citep{paper40} developed FishTrack, a multi-fish online tracking model with three branches: target detection, trajectory prediction, and re-identification. This model simultaneously establishes a fish motion model and an appearance model to achieve multi-fish online tracking. Mei et al. \citep{paper41} proposed a novel single-target fish tracking method, SiamFCA, which is based on a twin network and a coordinate attention mechanism. This method further enhances images using contrast-constrained adaptive histogram equalization to improve the accuracy and robustness of the model in complex scenes.

However, the aforementioned methods primarily address challenges such as occlusion and complex environments, while often overlooking the strong correlation between different parts of fish bodies and the temporal continuity between consecutive frames. This oversight leads to phenomena such as parts of fish bodies with similar shapes being mistakenly recognized as the same fish, as well as redundancy in the detection frames.

\subsection{Vision Mamba and RT-DETR}
In the field of visual representation learning, the introduction of the Vision Mamba \citep{paper42} model marks a significant advancement over traditional Transformer architectures. By leveraging the Bidirectional State Space Model (BSSM), Vision Mamba aims to overcome the scalability limitations of traditional self-attention mechanisms when handling long sequential data. Compared to the Vision Transformer \citep{paper43} model, the Vision Mamba, which processes image sequences through its BSSM mechanism, not only captures both global and local visual information but also achieves linear time complexity, significantly improving computational efficiency.

The Vision Mamba model's significant advantage lies in its ability to process long sequential data, enabling strong performance in vision tasks involving high-resolution images and video data. In recent years, numerous researchers have advanced this model by developing variations such as Fusion Mamba\citep{paper44}, MIM\citep{paper45}, and VMamba\citep{paper46}, establishing it as an emerging research direction in the field of computer vision.

DETR\citep{paper47}, which utilizes the Transformer architecture, is renowned for its capability to excel in accurate object localization and relationship modeling, as it transforms the target detection task into a multi-class classification problem, thereby making it particularly suitable for applications such as multi-object tracking. However, it must be noted that the model is characterized by high computational complexity, which necessitates a significant amount of computational resources. In contrast, the RT-DETR\citep{paper48} model, which combines the advantages of the Transformer architecture, is designed to significantly enhance both inference speed and accuracy. By efficiently integrating the encoder with IoU-aware query selection, it is able to eliminate the need for traditional NMS post-processing, a factor that contributes to its distinction as a breakthrough in the field of real-time target detection.

In the specialized field of target tracking, existing the Transformer-based frameworks encounter challenges related to target loss and computational inefficiency when tracking fast-moving objects over extended periods. To enhance tracking stability and accuracy, this study employs the Mamba in MIM framework in conjunction with RT-DETR, leveraging its efficient long-sequence processing capability and memory mechanism for feature extraction. This approach not only improves real-time target tracking but also increases adaptability to the motion characteristics of targets in complex environments, providing a novel technical solution for the realm of target tracking.

\section{Methods}
\subsection{Main Framework}
\begin{figure*}
  \centering
  \includegraphics[width=0.9\linewidth]{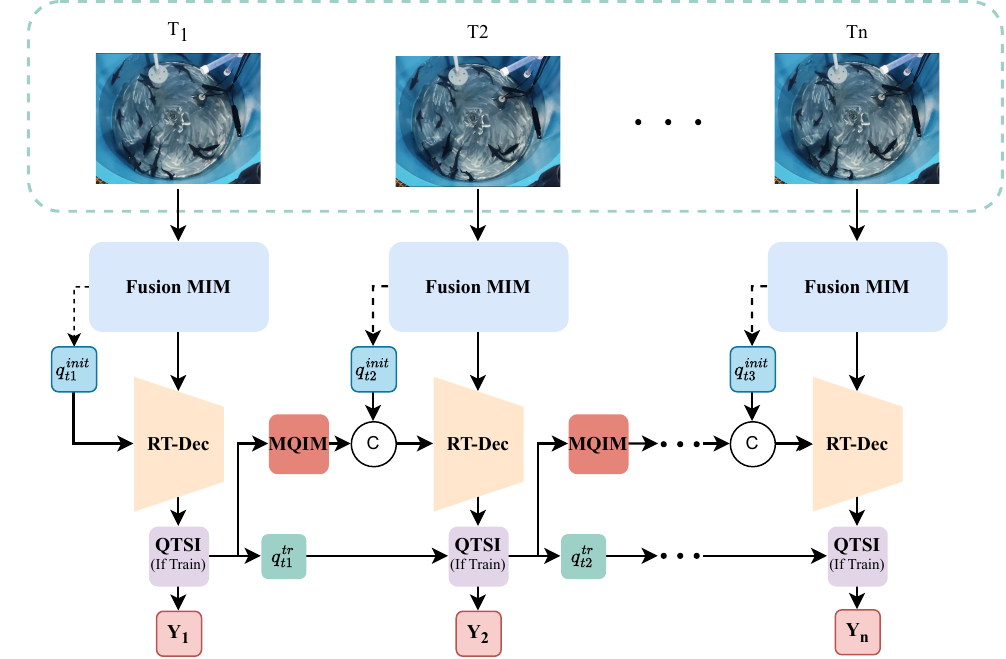}
  \caption{Main Framework diagram of the model.}
  \label{fig5}
  \vspace{-5px}
\end{figure*}

This framework leverages the MIM architecture for feature extraction, the RT-Decoder architecture for decoding, and the QTSI and MQIM modules for post-processing the detection and tracking queries. The overall structure is illustrated in Figure~\ref{fig5}. Initially, video sequences are sequentially input into the model, where each frame undergoes efficient feature extraction and encoding through the MIM architecture. To ensure the Query thoroughly learns the feature information, the model integrates the uncertainty minimal query selection unique to RT-DETR during the training phase. The initialized Detection Query is extracted from the encoder using RT-DETR's distinctive scheme and input into the RT-Decoder for decoding. For subsequent frames, the Track Query from the previous frame, combined with the decoded Detection Query, is used as the tracking query for the current frame after interaction with QTSI. Simultaneously, the Track Query and the newly initialized Detection Query are processed through MQIM to serve as the tracking query for the next frame.

\subsection{Fusion MIM}
To effectively extract fish boundary information in each frame, this paper incorporates the feature extraction mechanism of MIM \citep{paper45} into the feature extraction module. Although MIM offers the advantage of multi-scale feature representation, it faces challenges in enabling sufficient interaction between features at different scales. Deep features encapsulate rich detailed information, whereas shallow features are more adept at capturing global context. Existing feature fusion models, such as Feature Pyramid Network (FPN) \citep{paper49}, Debiased Single-Shot Multi-Frame Detector (DSSD) \citep{paper50}, Differential Typical Correlation Analysis \citep{paper51}, and Adaptive Spatial Feature Fusion (ASFF) \citep{paper52}, have made progress in inter-level feature correlation. However, these models still struggle with the detailed mapping of deeper features and the effective transfer of shallow global information.

\begin{figure*}
  \centering
  \includegraphics[width=0.9\linewidth]{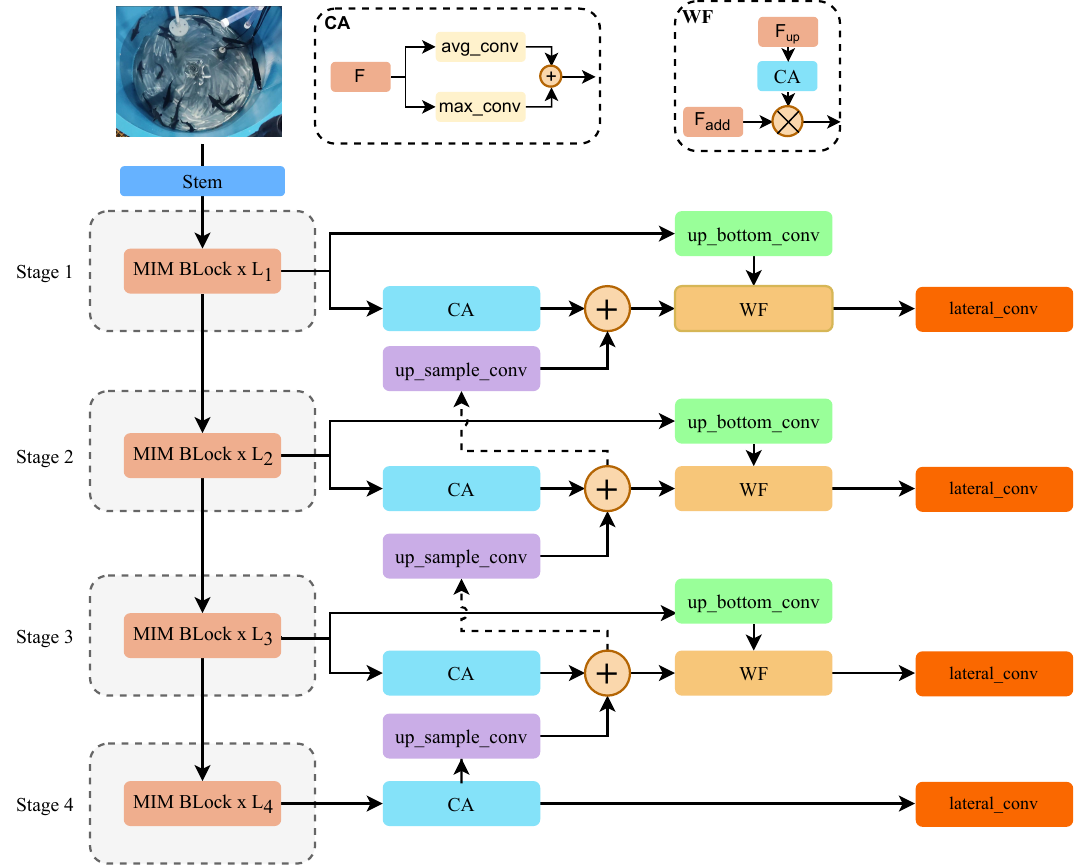}
  \caption{Fusion MIM Framework.}
  \label{fig6}
  \vspace{-5px}
\end{figure*}

To address this problem, this paper proposes an innovative feature extraction module, Fusion MIM, which combines the advantages of the Feature Fusion Single-Shot Multi-Frame Detector \citep{paper53}. This module is specifically designed to enhance the mapping of detailed information from deep features to shallow layers, while also improving the transfer of global information from shallow features. The Fusion MIM module is illustrated in Figure~\ref{fig6}.

This module employs four layers of MIM feature extraction sub-modules at different scales to obtain the initial visual word dimension $W_0\epsilon \mathbb{R} ^{\frac{H}{4} \times \frac{W}{4} \times C}$ and visual sentence dimension $S_0\epsilon \mathbb{R} ^{\frac{H}{4} \times \frac{W}{4} \times D}$ of the image by passing the video frame image $X\epsilon \mathbb{R} ^{H \times W \times 3}$ of the fish through the Stem module, as shown in Equation (1):
\begin{equation}
  W_0,S_0=Stem\left( X \right)
\end{equation}
Afterward, the initial visual word and visual sentence dimensions are passed through four MIM module stages of varying depths to obtain feature information at different scales $F$, as shown in Equations (2), (3), and (4):
\begin{equation}
  {W_l}^i,{S^i}_l={MIM^i}_l\left( {W^i}_{l-1},{S^i}_{l-1} \right),l=1,2,...,L_i
\end{equation}

\begin{equation}
  f_i=UpsampleBlock\left( S_{L_i} \right)
\end{equation}

\begin{equation}
  F=Cat\left( f_i \right),i=1,2,3,4
\end{equation}
Where $i$ represents the feature extraction process corresponding to the $i$-th layer, $L_i$ represents the depth of the MIM module for each layer, and the acquired features $F$ are fed into the feature fusion module for interactive information fusion, as shown in Equations (5) and (6):
\begin{equation}
  f_{enc}^{i}\begin{cases}
    WF\left[\begin{array}{c}
                UBConv\left( f_i \right),                                \\
                \left( \begin{array}{c}
                   USConv\left( DCA\left( f_{i+1} \right) \right) \\
                   +DCA\left( f_i \right)                         \\
                 \end{array} \right) \\
              \end{array} \right], i=1,2,3 \\
    DCA\left( f_i \right), i=4                                       \\
  \end{cases}
\end{equation}
\begin{equation}
  F_{enc}=Cat\left( f_{enc}^{i} \right), i=1,2,3,4
\end{equation}
Where $WF$ denotes the feature mapping module, designed to map deep detail information to shallow features through weighted fusion, and $DCA$ refers to the double cross-attention module, which captures long-range dependencies by sequentially addressing channel and spatial dependencies between multi-scale encoder features to bridge the semantic gap between encoder and decoder features. This well-designed feature fusion strategy effectively integrates various scale features in MIM, enhancing the richness and accuracy of feature representation and providing robust support for the precise extraction of the fish body boundary.

\subsection{Query Time Sequence Intersection}
\begin{figure}
  \centering
  \includegraphics[width=0.9\linewidth]{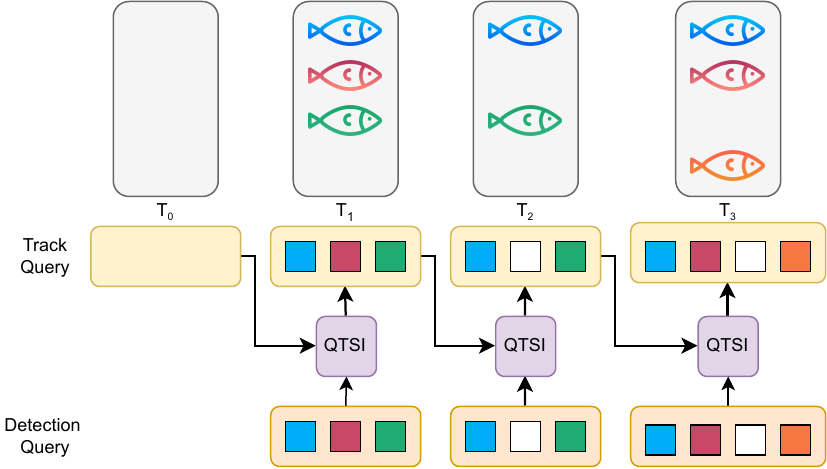}
  \caption{Query Time Sequence Intersection Framework.}
  \label{fig7}
  \vspace{-5px}
\end{figure}

Problems such as excessive similarity between individual sturgeon and severe occlusion among fish cause the original MOTR model overly rely on Track Query, leading to false tracking. To address this issue, this paper proposes a Query Temporal Interaction Module (QTSI) adapted from the MO-YOLO \citep{paper54} model. The QTSI enables the model to evenly distribute Query detection while minimizing additional computational burden. This module is utilized exclusively during the training phase. The main framework of the QTSI is illustrated in Figure~\ref{fig7}.

\begin{figure}
  \centering
  \includegraphics[width=0.9\linewidth]{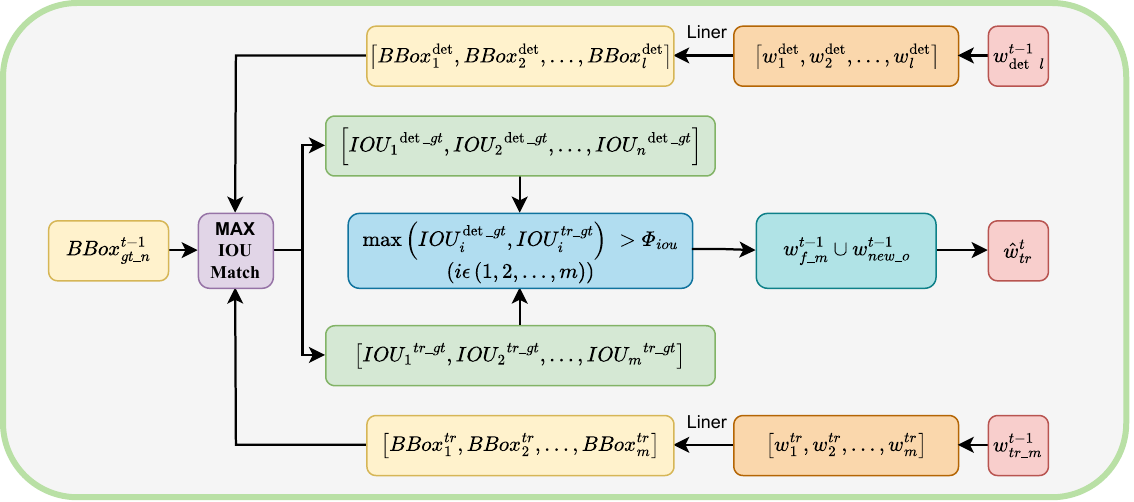}
  \caption{QTSI module calculation flowchart.}
  \label{fig8}
  \vspace{-5px}
\end{figure}

Since both Detect Query and Track Query contain the object's bounding box (BBOX) information, new objects are typically predicted through Detect Query. This approach helps avoid the issue in the original model where Track Query might incorrectly carry over from frame $t$ (where $t>2$) to subsequent frames for prediction. The specific calculation process of QTSI is shown in Figure~\ref{fig8}. Firstly, we define the following terms:

\begin{itemize}
  \item Detection Frame $BBox_{l}^{\det}$: Corresponds to the detection query $w_{\det \_l}^{t-1}$ from the previous frame.
  \item Tracking Frame $BBox_{m}^{tr}$: Corresponds to the tracking query $w_{tr\_m}^{t-1}$ from the previous frame, excluding new matches.
  \item Real Frame $w_{new\_o}^{t-1}$: The actual frame from the previous time step.
  \item Tracking Query $w_{new\_o}^{t-1}$: The tracking query from the previous frame, including new matches.
\end{itemize}

Here, $l$, $m$, $n$, and $o$ represent the number of detection frames, tracking frames, real frames, and new matching queries, respectively, with $m+o=n$. Define MIOU as the maximum IOU, ERF as each real frame, ETF as each tracking frame, and EDF as each detection frame. Then, to determine the maximum Intersection over Union (IOU) score:
\begin{enumerate}
  \itemsep=0pt
  \item Calculate the MIOU score of EDF with ERF, which is $\left[ {IOU_1}^{\det \_gt},{IOU_2}^{\det \_gt},...,{IOU_n}^{\det \_gt} \right]$.
  \item Calculate the MIOU score of ETF with ERF, which is $\left[ {IOU_1}^{tr\_gt},{IOU_2}^{tr\_gt},...,{IOU_m}^{tr\_gt} \right]$.
  \item Calculate the MIOU score of the above two results.
\end{enumerate}
If the maximum IOU score exceeds a predefined threshold $\varPhi _{iou}$, use the corresponding query $w_{f\_m}^{t-1}$ for further processing. Then, solve the concatenation result with $w_{new\_o}^{t-1}$ to determine the tracking query $w_{new\_o}^{t-1}$ for the subsequent frames, as shown in Equation (7).
\begin{equation}
  \hat{w}_{tr}^{t}=w_{f\_m}^{t-1}\cup w_{new\_o}^{t-1}
\end{equation}

\subsection{Mamba Query Interaction Module}
\begin{figure}
  \centering
  \includegraphics[width=0.9\linewidth]{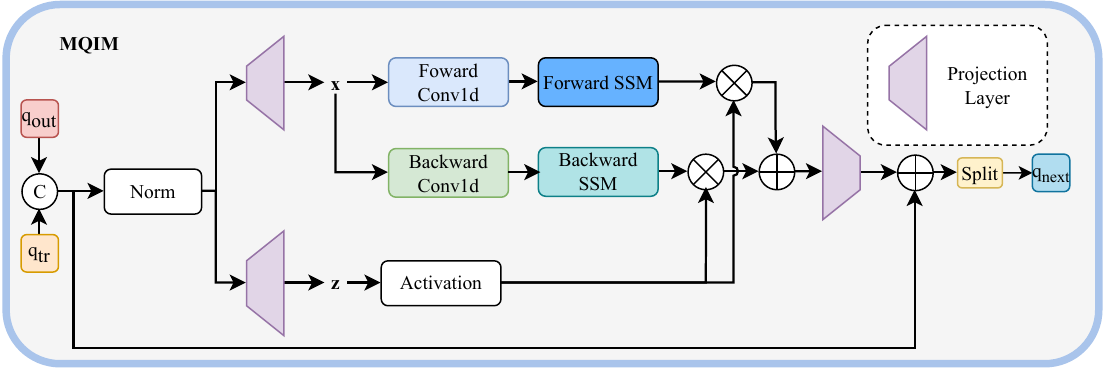}
  \caption{Mamba Query Interaction Module.}
  \label{fig9}
  \vspace{-5px}
\end{figure}

In this paper, we propose a new Mamba-based Query Fusion Interaction Module (MQIM), which transforms the original QIM module into a Mamba-based temporal feature interaction module. By leveraging the bidirectional temporal interaction mechanism of Vision Mamba \citep{paper42}, MQIM facilitates feature association across multiple frames through long-term feature memory and feedback. This enhancement improves the tracking of fast-moving objects. The module diagram is shown in Figure~\ref{fig9}.

By interacting the tracking query processed by the QTSI module with the corresponding decoder output for temporal features, the initial tracking query of the next frame gains prior knowledge from the previous frame. This process enables better generalization of the previous frame's tracking results to the tracking of the next frame.

\subsection{Joint Average Loss}
In this paper, multiple loss modules are utilized for optimization, with the loss for a single frame image calculated as shown in Equation (8).

\begin{equation}
  \mathcal{L} \left( \left. \hat{Y}_i \right|_{\omega _i},Y_i \right) =\lambda _{cls}\mathcal{L} _{cls}+\lambda _{l_1}\mathcal{L} _{l_1}+\lambda _{\mathrm{giou }}\mathcal{L} _{\mathrm{giou }}
\end{equation}
where $\mathcal{L} _{cls}$ is the focal loss \citep{paper55}, $\mathcal{L} _{l_1}$ is the L1 loss, and $\mathcal{L} _{giou}$ is the GIOU loss \citep{paper56}. $\lambda _{cls}, \lambda _{l_1}\,\,and\,\,\lambda _{giou}$ are the corresponding weight coefficients.

Since the MOTR algorithm unifies the common loss of multi-frame images as the overall loss for the entire video sequence, effectively improving the tracking performance of time-sequenced video sequences, this paper also adopts this approach to optimize the loss calculation. The track loss and detector loss are calculated according to Equation (8), then summed and averaged as shown in Equation (9) and (10):

\begin{equation}
  M=\sum_{n=1}^N{\left( \mathcal{L} \left( \left. \hat{Y}_{i}^{tr} \right|_{\omega _{i}^{tr}},Y_{i}^{tr} \right) +\mathcal{L} \left( \left. \hat{Y}_{i}^{\det} \right|_{\omega _{i}^{\det}},Y_{i}^{\det} \right) \right)}
\end{equation}

\begin{equation}
  \mathcal{L} _o\left( \left. \hat{Y} \right|_{\omega},Y \right) =\frac{M}{\sum_{n=1}^N{V_i}}
\end{equation}
Where $V_i=V_{i}^{tr}+V_{i}^{\det}$ represents the total number of ground truths in frame $i$, $\mathcal{L} \left( \left. \hat{Y}_{i}^{tr} \right|_{\omega _{i}^{tr}},Y_{i}^{tr} \right)$ and $\mathcal{L} \left( \left. \hat{Y}_{i}^{\det} \right|_{\omega _{i}^{\det}},Y_{i}^{\det} \right)$ represent the tracking loss and detection loss of frame $i$, respectively.

\section{Experiments}
\subsection{Datasets and Settings}
\begin{figure}
  \centering
  \includegraphics[width=0.9\linewidth]{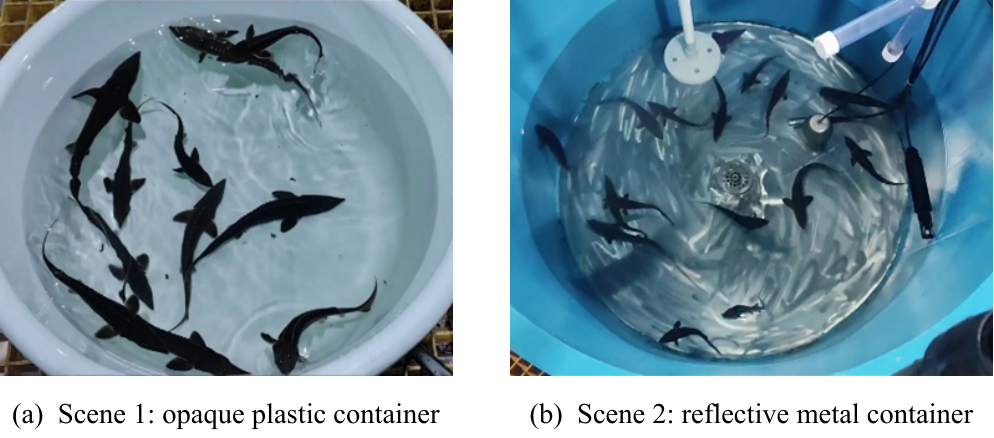}
  \caption{Two different data collection environments.}
  \label{fig10}
  \vspace{-5px}
\end{figure}

In this paper, a new sturgeon fish tracking dataset was constructed. The video clips in this dataset were collected at the National Innovation Center for Digital Fishery of China Agricultural University. The camera resolution was set to 1920 × 1080, with a frame rate of 30 FPS. The video clips were captured in two different experimental scenarios, as shown in Figure~\ref{fig10}.

To simulate a realistic aquaculture scenario (e.g., environments involving bubbles, water rotation, etc.), Scene 2 was added to water circulation and oxygenation devices, as well as sensors of various morphologies. The video attributes are detailed in Table~\ref{tab:tab1}, with video clip lengths ranging from 10 to 20 seconds. A total of 11,000 labeled video frames were obtained through video frame-splitting and automatic labeling techniques. Specific dataset information, including the division of training and testing datasets, is presented in Table~\ref{tab:tab1}.

\begin{table*}[!h]
  \centering
  \caption{Dataset description('both' means all cases are included).}
  \setlength{\tabcolsep}{14.5pt}
  \label{tab:tab1}
  \begin{tabular}{lllllll} \toprule
    Scenes                   & Have Sensors & \begin{tabular}[l]{@{}l@{}}Have Water circulation \\ and aeration units\end{tabular} & illumination & Videos & Frames & Type  \\ \midrule
    \multirow{3}{*}{Scene 1} & no           & no                                                                                   & light        & 4      & 1700   & train \\
                             & no           & no                                                                                   & common       & 4      & 1700   & train \\
                             & no           & no                                                                                   & both         & 3      & 600    & test  \\ \hline
    \multirow{3}{*}{Scene 2} & yes          & no                                                                                   & both         & 4      & 1500   & train \\
                             & yes          & yes                                                                                  & both         & 4      & 1500   & train \\
                             & yes          & both                                                                                 & both         & 3      & 1000   & test  \\  \bottomrule
  \end{tabular}
\end{table*}

The hardware environment for this experiment includes CPU: 13th Gen Intel® Core™ i9-13900K x 32, RAM: 128 GB, and GPU: 2 x NVIDIA GeForce RTX™ 4090. The operating system is Ubuntu 23.04, and the code is implemented using the PyTorch framework. Other relevant experimental parameters are detailed in Table~\ref{tab:tab2}.

\begin{table}[!h]
  \caption{Experiment Setting.}
  \setlength{\tabcolsep}{25pt}
  \label{tab:tab2}
  \begin{tabular}{ll} \toprule
    Experimental parameters & Value          \\ \midrule
    batch\_size             & 1              \\
    epoch                   & 100            \\
    loss\_rate              & $2x10^{-4}$    \\
    sampler\_steps          & {[}20,50,70{]} \\
    sampler\_lengths        & {[}2,3,4,5{]}  \\
    miss\_tolerance         & 100            \\
    \bottomrule
  \end{tabular}
\end{table}

\subsection{Evaluation Metrics}

To demonstrate the superiority of the proposed FMRFT model, it is evaluated using several metrics, including Multi-Object Tracking Accuracy (MOTA), Multiple Object Tracking Precision (MOTP) \citep{paper57,paper58}, Identification F1-score ($IDF_1$) \citep{paper59}, Identification Precision (IDP), Identification Recall (IDR), Frames Per Second (FPS), and Training Memory Allocation (TMA) per GPU.

MOTA measures the accuracy of single-camera multi-target tracking, expressed by the Equation (11):
\begin{equation}
  MOTA=1-\frac{FN+FP+\Phi}{T}
\end{equation}
Where $FN$ is the miss rate (i.e., positive samples predicted as negative by the model), $FP$ is the false alarm rate (i.e., negative samples predicted as positive by the model), $\varPhi$ represents the sum of target jumps across all frames (i.e., changes in the tracking trajectory from "tracking" to "no-tracking"), and $T$ is the total number of true targets in all frames. The closer MOTA is to 1, the better the performance of the tracker.

MOTP is a measure of the accuracy of single-camera multi-target tracking matching, which refers to the distance between the predicted trajectory and the true trajectory, reflecting the accuracy of the tracking results, and is expressed by the Equation (12):
\begin{equation}
  \mathrm{MOTP}=\frac{\sum_{i,t}{d_{t}^{i}}}{\sum_t{c_t}}
\end{equation}
Where $c_t$ denotes the number of matches in frame $t$. The matching error is computed for each pair of matches, and $d_{t}^{i}$ is the bounding box overlap between hypothesis $i$ and its assigned ground truth object.

$IDF_1$ refers to the F1 score for object ID identification in each object frame, which calculated using the Equation (13) shown below:
\begin{equation}
  IDF_1=\frac{2IDTP}{2IDTP+IDFP+IDFN}
\end{equation}
Furthermore, $IDP$ and $IDR$ are used to evaluate the performance of detector and tracker in more detail. The formulas for calculating $IDP$ and $IDR$ are shown in Equations (14) and (15):
\begin{equation}
  IDP=\frac{IDTP}{IDTP+IDFP}
\end{equation}
\begin{equation}
  IDR=\frac{IDTP}{IDTP+IDFN}
\end{equation}
Where $IDTP$ and $IDFP$ represent the number of true positive IDs and false positive IDs, respectively, while IDFN represents the number of false negative IDs. $IDTP$ is the sum of the weights of the edges selected as true positive ID matches, indicating the percentage of correctly assigned detections throughout the entire video. $IDFN$ represents the total weight of the selected false negative ID edges, while $IDFP$ denotes the total weight of the selected false positive ID edges.

\subsection{Experimental Evaluation}
To highlight the superiority of the models proposed in this paper, we trained and tested the mainstream Tracking by Detection and Tracking by Query methods on the newly introduced sturgeon dataset using the FMRFT multi-target fish tracking algorithm, and all under the same experimental conditions. The experimental results are presented in Table~\ref{tab:tab3}. In this table, OC-SORT \citep{paper29} and FairMOT \citep{paper60} are detection-based models, while the remaining models are query-based, which include TrackFormer \citep{paper36}, TransCenter \citep{paper09}, MOTR \citep{paper10}, MOTIP \citep{paper62}, and FMRFT (proposed in this paper).

\begin{table*}[!h]
  \caption{Results of the comparison with state-of-the-art MOT models(with ↓ indicating lower values are better, ↑ indicating higher values are better).}
  \setlength{\tabcolsep}{10.5pt}
  \label{tab:tab3}
  \begin{tabular}{lllllllll} \toprule
    Category                                                                         & Model       & $IDF_1$↑         & IDP↑             & IDR↑             & MOTA↑            & MOTP↓          & FPS↑         & TMA↓        \\ \midrule
    \multirow{2}{*}{\begin{tabular}[l]{@{}l@{}}Tracking by\\ Detection\end{tabular}} & OC-SORT     & 56.00\%          & 48.00\%          & 67.20\%          & 34.40\%          & 0.858          & \textbf{116} & 23G         \\
                                                                                     & FairMOT     & 49.40\%          & 59.90\%          & 42.00\%          & 59.40\%          & 0.241          & 73           & 20G         \\ \hline
    \multirow{5}{*}{\begin{tabular}[l]{@{}l@{}}Tracking by\\ Query\end{tabular}}     & TrackFormer & 76.30\%          & 76.40\%          & 76.10\%          & 92.30\%          & 0.103          & 21           & 15G         \\
                                                                                     & TransCenter & 60.20\%          & \textbf{90.30\%} & 45.10\%          & 48.80\%          & 0.267          & 63           & 8G          \\
                                                                                     & MOTR        & 64.90\%          & 59.00\%          & 72.00\%          & 61.00\%          & \textbf{0.091} & 46           & 10G         \\
                                                                                     & MOTIP       & 87.50\%          & 87.70\%          & 87.40\%          & 91.70\%          & 0.903          & 52           & 12G         \\
                                                                                     & FMRFT(ours) & \textbf{90.30\%} & 90.10\%          & \textbf{90.40\%} & \textbf{94.30\%} & 0.123          & 50           & \textbf{6G} \\
    \bottomrule
  \end{tabular}
\end{table*}

From Table~\ref{tab:tab3}, it can be observed that FMRFT achieves the highest, $IDF_1$, IDR, and MOTA scores, with values of 90.3\%, 90.4\%, and 96.3\%, respectively, while also maintaining a lower MOTP of 0.123. Additionally, FMRFT demonstrates a good FPS and low video memory usage during training.

Compared to traditional detection-based models, FMRFT exhibits significant advantages in multi-target fish tracking tasks. Although detection-based models excel in detection speed, they fall short in tracking and recognition accuracy, particularly in complex scenes. In contrast, query-based multi-target tracking models show improvements in relevant metrics. Among these, TransCenter achieves the highest IDP, and MOTR achieves the lowest MOTP. However, FMRFT delivers the best overall performance, especially in IDF1 and MOTA metrics.

\begin{figure*}
  \centering
  \includegraphics[width=0.9\linewidth]{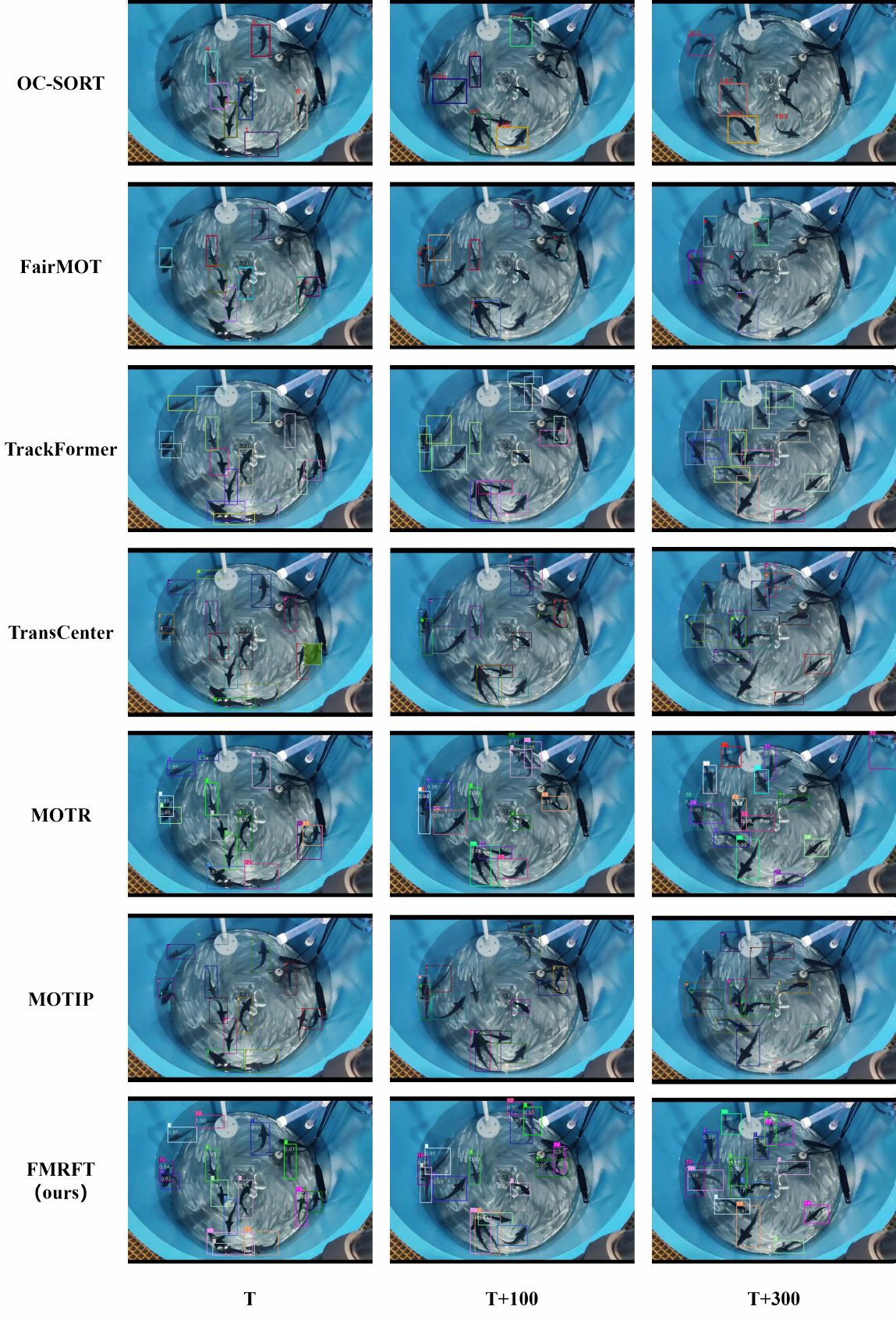}
  \caption{Tracking Results of Different Models at T, T+100, and T+300 Moments (T=0, in Frames Per Second).}
  \label{fig11}
  \vspace{-5px}
\end{figure*}

To further validate the effectiveness of the FMRFT model, we demonstrated its performance in the same experimental scenario by visualizing the tracking results. As shown in Figure~\ref{fig11}, comparing the tracking results at moments T, T+100, and T+300, the detection-based model performs well in object tracking, but a large number of non-detections occur when multiple objects are tracked simultaneously. For the query-based TransCenter and TrackFormer models, the tracking effectiveness decreases significantly when there is a large area of fish occlusion. Although MOTR and MOTIP perform well in recognition and tracking, there are more redundant frames, with this issue being especially pronounced in long-term tracking. In contrast, the FMRFT model proposed in this paper exhibits robust tracking performance over extended periods, even in complex scenarios such as occlusion, strong illumination, and sensor interference.

\subsection{Ablation Experiment}
In the proposed FMRFT model, the Fusion MIM architecture is utilized for feature extraction, while the RT-Decoder architecture is employed for encoding, enabling accurate identification of multi-target fish with enhanced robustness in complex environments, such as occlusion and glare. The QTSI module is then applied for post-processing detection and tracking queries, effectively minimizing issues related to multiple labels being assigned to the same tracking object. Additionally, the MQIM facilitates feature memory and association across multiple frames, thereby improving tracking performance for fast-moving objects. To validate the effectiveness of each module, ablation experiments were conducted, with results presented in Table~\ref{tab:tab4}. These results shows that incorporating the MQIM, Fusion MIM, and QTSI modules into the main framework enhances all performance metrics, with a 7.8\% increase in IDF1, an 8.8\% increase in MOTA, and a 0.048 decrease in MOTP compared to the main framework model. This further confirms the effectiveness of the individual modules.

\begin{table*}[!h]
  \caption{Effectiveness of different components of FMRFT(with ↓ indicating lower values are better, ↑ indicating higher values are better).}
  \setlength{\tabcolsep}{15.5pt}
  \label{tab:tab4}
  \begin{tabular}{llllllll} \toprule
    Sequence & \begin{tabular}[l]{@{}l@{}}Main\\ Framework\end{tabular} & MQIM                & \begin{tabular}[l]{@{}l@{}}Fusion\\ MIM\end{tabular} & QTSI                & IDF1↑            & MOTA↑            & MOTP↓          \\ \midrule
    1        & \textbf{\checkmark}                                      & \textbf{}           & \textbf{}                                            & \textbf{}           & 70.50\%          & 73.50\%          & 0.171          \\
    2        & \textbf{\checkmark}                                      & \textbf{\checkmark} & \textbf{}                                            & \textbf{}           & 85.20\%          & 87.30\%          & 0.125          \\
    3        & \textbf{\checkmark}                                      & \textbf{\checkmark} & \textbf{\checkmark}                                  & \textbf{}           & 86.40\%          & 92.80\%          & 0.161          \\
    4        & \textbf{\checkmark}                                      & \textbf{\checkmark} & \textbf{\checkmark}                                  & \textbf{\checkmark} & \textbf{90.30\%} & \textbf{94.30\%} & \textbf{0.123} \\
    \bottomrule
  \end{tabular}
\end{table*}

\begin{figure*}
  \centering
  \includegraphics[width=0.9\linewidth]{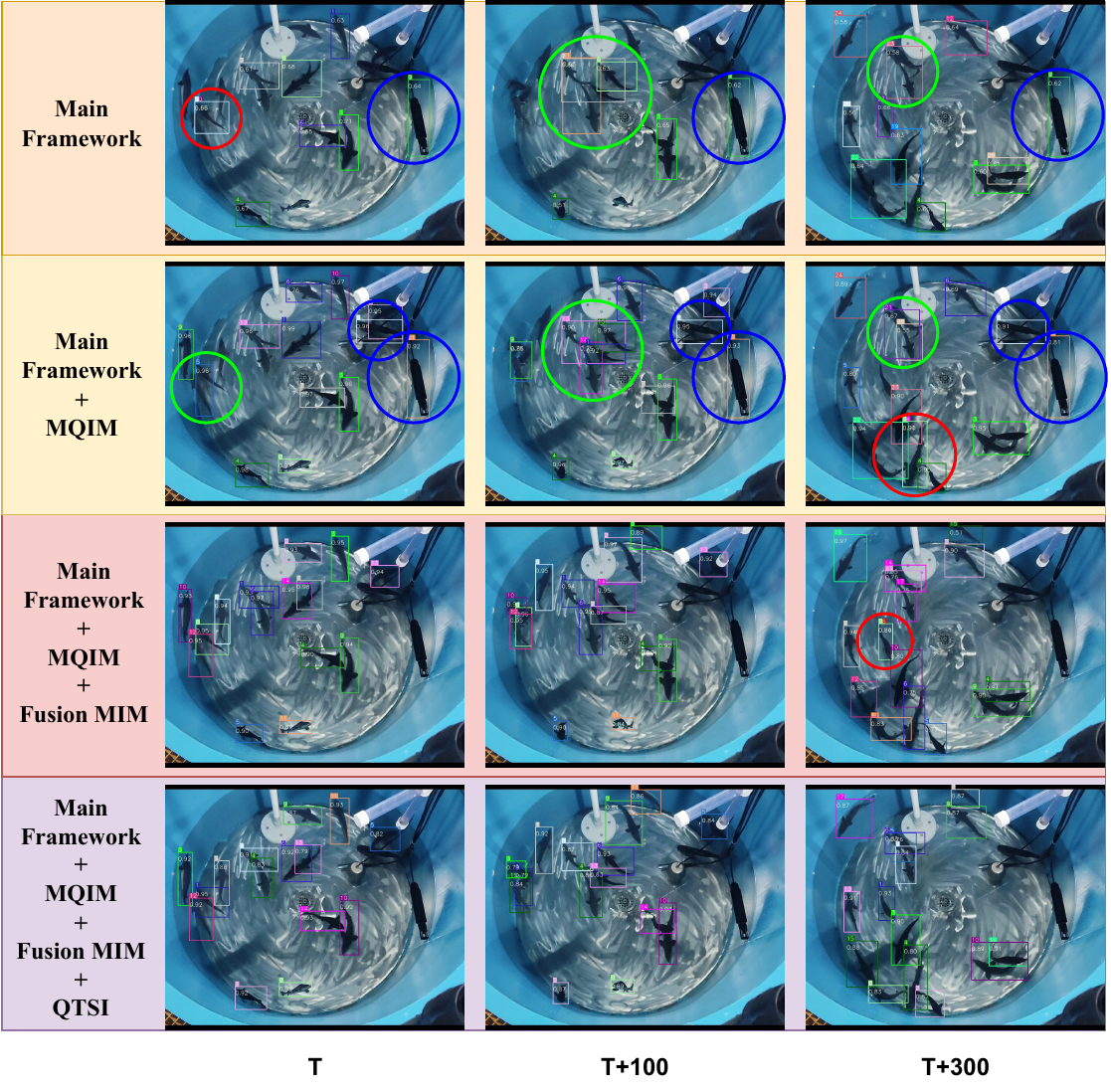}
  \caption{Tracking Results of Different Module Combinations at T, T+100, and T+300 Moments (T=0, in Frames Per Second).}
  \label{fig12}
  \vspace{-5px}
\end{figure*}

To further demonstrate the role of each module, Figure~\ref{fig12} illustrates that, although the main frame alone achieves a good tracking effect, it also exhibits misdetections and omissions, indicated by the green circle in the figure. Additionally, sensor interference from other devices results in incorrect detections, as shown by the blue circle. However, with the progressive integration of the relevant modules, the occurrences of misdetection and omission are gradually reduced, significantly improving object tracking accuracy.

In the third row of Figure~\ref{fig12}, when the OTSI module is not used within the FMRFT model, redundant tracking frames for a single object can still arise during extended tracking periods, as exemplified by the red frame in the last image. Conversely, in the last row of the figure, the introduction of the OTSI module addresses this issue by assigning higher weights to newly detected objects, thereby mitigating the occurrence of redundant frames caused by excessive reliance on the tracking results from previous frames. Furthermore, under the combined influence of the MQIM and Fusion MIM modules, the tracking results demonstrate notable improvements.

\subsection{Visualization and Generalizability}

\begin{figure}
  \centering
  \includegraphics[width=0.9\linewidth]{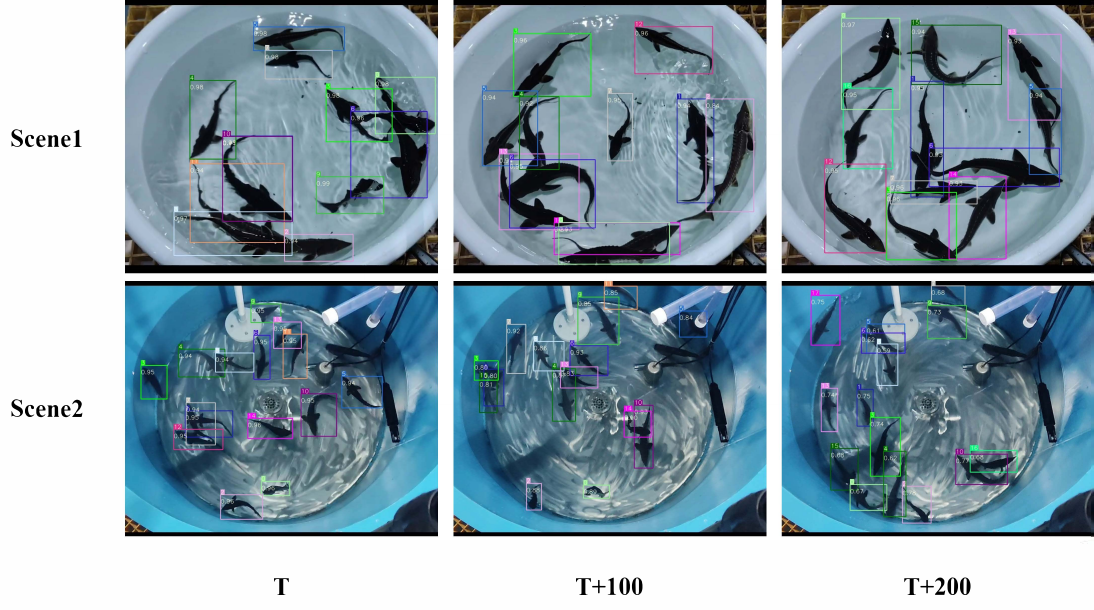}
  \caption{Tracking Results in Different Experimental Scenarios at T, T+100, and T+200 Moments (T=0, in Frames Per Second).}
  \label{fig13}
  \vspace{-5px}
\end{figure}

To visualize the tracking accuracy of the FMRFT model across different experimental scenarios, this section applies the model to the scenes depicted in Figure~\ref{fig13}, further validating its robustness. In Scene 1, despite the relatively simple environment characterized by overlapping occlusions and splashing—due to minimal morphological differences among the sturgeons—the model demonstrated excellent tracking performance. In Scene 2, FMRFT maintained high performance even amidst challenges such as reflections from metal containers, water circulation, and interference from sensors and oxygenation equipment. It accurately tracked each sturgeon from time T to T+200, showcasing high recognition accuracy despite the complex environmental factors.

\begin{figure}
  \centering
  \includegraphics[width=0.9\linewidth]{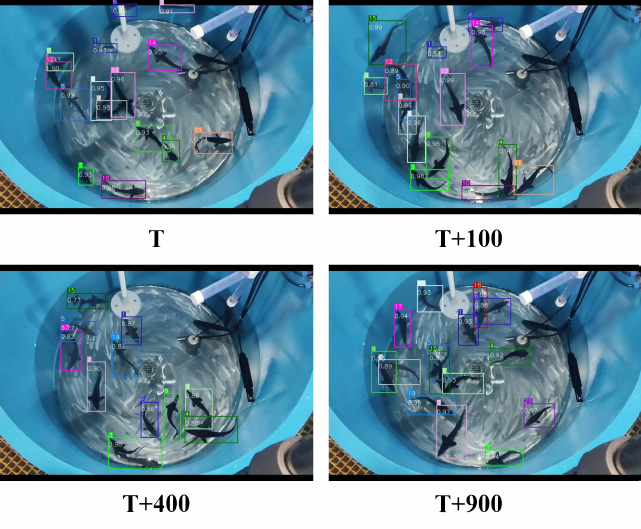}
  \caption{Tracking Results in Scene 2 at T, T+100, T+400, and T+900 Moments (T=0, in Frames Per Second).}
  \label{fig14}
  \vspace{-5px}
\end{figure}

To verify the long-term stability of the model, the tracking performance of FMRFT was evaluated at moments T, T+100, T+400, and T+900 in Scene 2, as shown in Figure~\ref{fig14}. The results indicate that FMRFT maintains consistent and stable tracking over time. However, some sturgeon was either excessively occluded or exited the experimental scene, suggesting that new tracking IDs might need to be assigned. To further optimize the model, future work will focus on integrating methods such as dedicated feature memory to enhance the tracking stability of objects that disappear from the scene.

\section{Conclusion}
In this paper, we propose a real-time fish tracking model, FMRFT, based on query-timing interaction which integrates the MIM and RT-DETR frameworks to address the issues of occlusion and redundant detection frames in complex scenes. 
The model adopts a fusion framework of MIM and RT-DETR with low memory footprint. Given the high similarity among fish and the presence of interfering objects in the scene, we introduced a new feature extraction framework Fusion MIM for designed for in-depth extraction of fish features. To mitigate the problem of multiple overlapping redundant tracking frames for a single fish, we developed a novel query-timing interaction module and an MIM-based feature interaction module to improve tracking correlation between consecutive frames and remove redundant frames. Tested on the newly proposed sturgeon fish tracking dataset, the model achieved 90.3\% $IDF_1$ and 94.3\% MOTA, demonstrating its effectiveness. Ablation experiments confirmed that the model maintains accurate and stable fish tracking performance under varying conditions, including bright light, reflections, and water waves. Overall, the proposed fish tracking model is well-suited for complex scenarios and offers a new solution for fish tracking in factory aquaculture. Future research will focus on further optimizing the model to address issues such as tracking ID reassignment due to excessive occlusion and re-entry of objects into the scene. Additionally, the dataset will be expanded, and new experimental scenarios will be introduced to meet a broader range of application requirements.

\section*{Acknowledgements}
This research was supported by the Mandarin fish factory farming service project(202305510811525), the Research and creation of key technologies for digital fishery intelligent equipment(2021TZXD006), the Intelligent aquaculture service project of water purification fish(202305510810180) and the Automatic counting software system for factory breeding standard fry(202305410811526).




\printcredits

\bibliographystyle{cas-model2-names}

\bibliography{cas-refs}



\end{document}